\pdfoutput=1
\documentclass{article}
\usepackage{iclr2027_conference,times}

\usepackage{amsmath,amsfonts,bm}

\def\eqref#1{equation~\ref{#1}}

\def\1{\bm{1}}

\DeclareMathAlphabet{\mathsfit}{\encodingdefault}{\sfdefault}{m}{sl}
\SetMathAlphabet{\mathsfit}{bold}{\encodingdefault}{\sfdefault}{bx}{n}

\usepackage{footmisc}
\usepackage{marvosym}
\usepackage{hyperref}
\usepackage{url}
\usepackage{booktabs}
\usepackage{graphicx}
\usepackage{float}
\usepackage{xcolor}

\DefineFNsymbols{lettersymbol}{\Letter \dagger \ddagger \S \P | {**}}
\setfnsymbol{lettersymbol}

\newcommand{\method}{\textsc{Oneira}}
\newcommand{\todo}[1]{\textcolor{red}{[TODO]}}
\newcommand{\num}[1]{#1}

\title{\method: From Open-Ended Generation to Open-World Interaction in Video World Models}

\author{\\
\textbf{Xindi Yang$^{1}$\quad Baolu Li$^{3}$\quad Liam Lee$^{4}$\quad Zhenfei Yin$^{5}$\quad Songxin Zhang$^{4}$\quad Zhuoyang Song$^{4}$} \\
\textbf{Xu Jia$^{3}$\quad Jianfei Cai$^{1}$\quad Tien-Tsin Wong$^{1}$\quad Bingyi Jing$^{4}$\quad Mengyue Yang$^{2}$\thanks{Corresponding author.}} \\
\\
$^1$Monash University\quad $^2$University of Bristol\quad $^3$Dalian University of Technology \\
$^4$The Chinese University of Hong Kong, Shenzhen\quad $^5$Oxford University\\
}

\iclrfinalcopy
\begin{document}

\maketitle
\lhead{}
\renewcommand{\headrulewidth}{0pt}

\begin{figure}[H]
\vspace{-7mm}
\centering
\includegraphics[width=\linewidth]{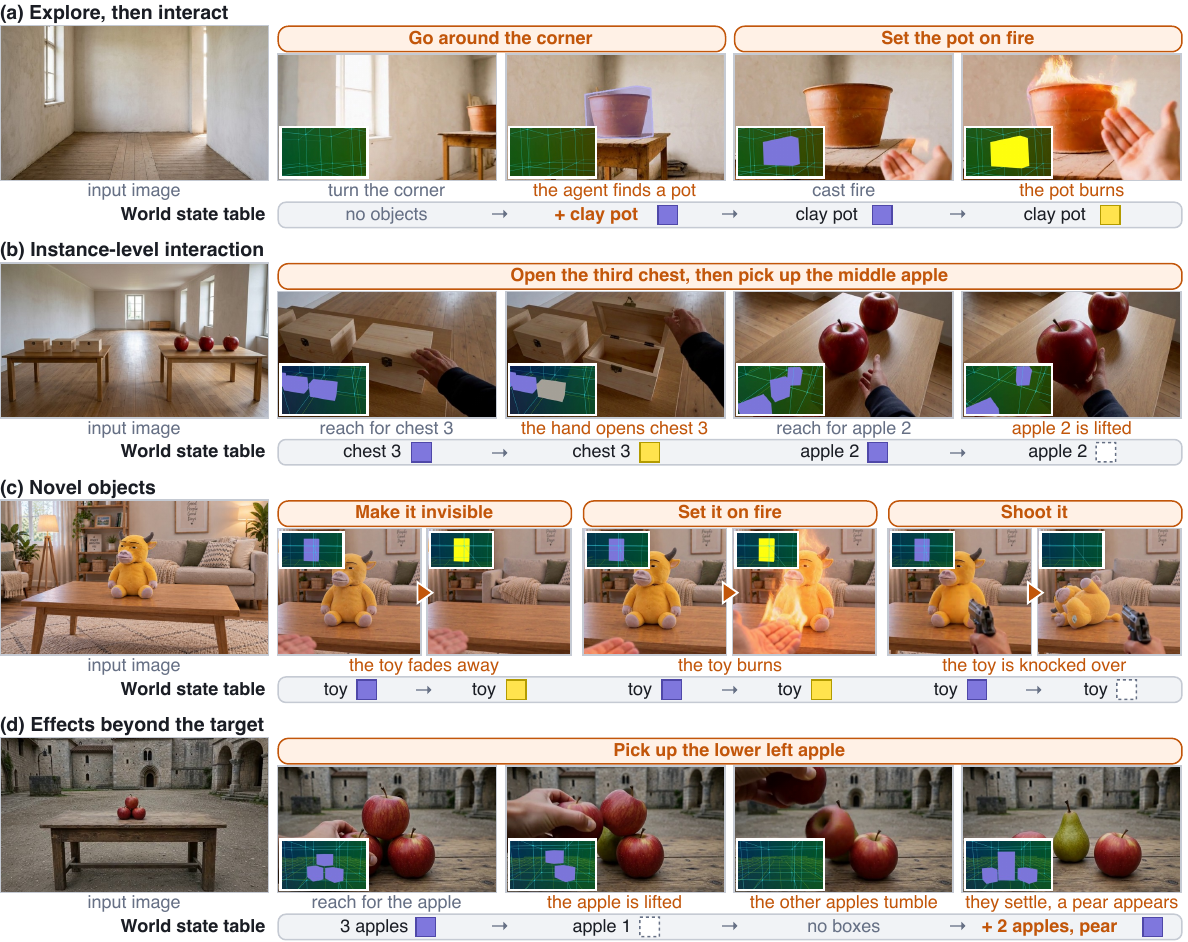}
\caption{\method{} conditions video generation on an explicit world state table that a coding agent updates.
Given an input image and a goal, the agent plans each interaction and writes its outcome into the table, and the video model renders the updated state into realistic video.
This design enables navigation and interaction beyond the input image (a), instance-level control (b), generalization to novel objects (c), and effects beyond the target object (d).
The small image in the corner of each frame shows the coarse conditioning video rendered from the table.
Purple marks the default state, yellow a changed state, and a dashed outline a box removed from the table.}
\label{fig:teaser}
\end{figure}

\begin{abstract}
Generative video world models can now synthesize open-ended environments that agents can navigate and interact with in simple ways.
Yet \emph{open-ended generation does not imply full interaction}: as a generated world expands, newly created content through navigation should expand what the agent can act upon, and as the agent changes the world, those changes should become persistent parts of the environment rather than transient visual effects.
We characterize these two requirements as \textbf{Open-World Interactivity}, where newly generated or encountered entities are incorporated into the actionable world, and \textbf{Persistent State}, where interaction outcomes are committed to the world state and continue to influence subsequent observations and interactions.
We present \textbf{\method{}}, an interactive video world model that closes the loop between generation and interaction through an explicit, extensible world state managed by a coding agent.
Given the current observation and an action or high-level goal, the agent reads the world state, grounds the relevant entities, plans the interaction, and writes its outcome back into a world state table.
When exploration reveals new objects, the agent incorporates them from generated observations, allowing the interaction space to expand with the generated world.
Meanwhile, previously induced state changes are carried across video segments, making the consequences of interaction persistent parts of subsequent world evolution.
The updated world state is rendered along the camera action trajectory into a coarse conditioning video, from which a video generator fills in the appearance, motion, and interaction details not represented in the state.
Experiments show that \method{} enables direct and consistent interaction with newly generated objects, while preserving the effects of prior interactions over long horizons.
Project page: \url{https://madaoer.github.io/projects/oneira/}
\end{abstract}

\section{Introduction}

Generative video world models can now produce increasingly rich environments that agents can explore, control, and interact with \citep{valevski2025diffusion,decart2024oasis,yang2023learning,parker2025genie}.
As these models move beyond fixed, preconstructed environments, however, a new question arises: \emph{how should interaction itself evolve when the world is being generated on the fly?}
In a conventional simulator or game engine, an object that exists in the environment is represented in an authoritative world state, and an action modifies that state.
In a generative world, by contrast, newly synthesized content may exist only in the generated observations, while the effects of an action may remain only in recent visual context.
The world can therefore expand visually without its space of possible interaction expanding with it.

We argue that \emph{open-ended generation does not imply full interaction}.
A fully interactive generative world requires two properties.
First, \textbf{Open-World Interactivity}: as exploration reveals or generates new entities, the interaction space should expand with the world, so that newly created content can become explicit targets of subsequent actions.
Second, \textbf{Persistent State}: when an agent changes the world, the outcome of that interaction should become a durable part of the world state and continue to constrain future observations and interactions.
In other words, what appears in the generated world should become actionable, and what the agent changes should remain changed.

Achieving both properties is difficult because existing approaches make different trade-offs.
Generative video world models have rapidly expanded the range of possible controls, from camera navigation and player inputs to robot actions, language instructions, and text-specified world events \citep{sun2025worldplay,hong2025relic,xu2026wonder,zhu2026sana, valevski2025diffusion,decart2024oasis,yang2023learning, parker2025genie,xiang2025pan}.
Persistent memory can further improve long-horizon consistency \citep{xiong2026actworld}.
However, the world state in these models is still largely implicit in generated pixels, prompts, or latent context.
As a result, a newly generated object is not necessarily registered as an actionable world entity, and a visually plausible interaction outcome is not necessarily committed as persistent state.
Explicit 3D worlds and conventional simulators provide the opposite strength: objects and their states can be referenced, modified, and preserved reliably, but the interactable world is largely determined by preconstructed geometry, assets, and interaction logic.
The missing capability is therefore not merely explicit state, but an \emph{explicit state that can itself grow with the generated world}.

We present \method{}, an interactive video world model that addresses this gap by separating \emph{world evolution} from \emph{visual realization}.
\method{} maintains an explicit, extensible world state table that is continuously managed by a coding agent.
Given the current observation and an action or high-level goal, the agent reads the world state, grounds the relevant entities, plans the interaction, and writes its outcome back into the table.
When exploration reveals new objects, the agent can incorporate them from generated observations into the world state, allowing the interaction space to expand beyond the content visible in the initial scene.
Conversely, when an interaction changes an existing object, the resulting state is explicitly recorded and carried forward across subsequent video segments.
The updated state is then rendered along the planned camera trajectory into a coarse conditioning video, while a video generator fills in the appearance, motion, and interaction details that are not explicitly represented in the state.
Thus, the explicit world state determines \emph{what exists and what changes}, while the video model determines \emph{how those changes look and move}.

This design closes the loop between generation and interaction: newly generated content can be grounded into actionable state, interaction outcomes are committed back into persistent interactive world state, and the updated state in turn conditions future generation.
As shown in Fig.~\ref{fig:teaser}, \method{} enables direct, instance-level interaction with objects that emerge during open-world exploration, while preserving the consequences of previous interactions across long video horizons.
Because the interaction state is represented independently of object appearance and category, the same mechanism can further generalize interaction control to object categories unseen during fine-tuning.

Our contributions are threefold.
(1) We introduce \method{}, an interactive video world model that enables \textbf{open-world interactivity} and \textbf{persistent state} through an explicit, extensible world state continuously updated by a coding agent.
(2) We design a world state table and a coarse conditioning video as the interface between explicit world evolution and video generation, allowing instance-level interaction with newly generated objects while preserving interaction outcomes across segments.
(3) We further construct a training pipeline that aligns videos with camera paths, world state tables, and chunk-level interaction captions, using the same state representation at training and inference.
\vspace{0mm}
\section{Related Work}

\method{} consists of a coding agent that maintains an explicit world state and a video model that renders observations from it.
We organize related work around interactive world models, explicit state representations, and structured video conditioning.
A more detailed review is in Appendix~\ref{app:relatedwork}.

\textbf{Interactive video world models.}
Interactive video world models predict visual evolution under agent actions, spanning game controls, camera navigation, robot actions, and semantic events \citep{valevski2025diffusion,alonso2024diffusion,yang2023learning,parker2025genie,sun2025worldplay,xiong2026actworld,gao2026infinite}.
Recent systems improve long-horizon consistency through memory and support increasingly rich interaction.
However, their primary modeling object remains the generated observation sequence, with interaction outcomes stored implicitly in pixels, latent context, or memory.
In contrast, \method{} writes each interaction outcome into an explicit world state before generation and uses the video model only to render the resulting state.

\textbf{Explicit world state.}
World models may represent dynamics in latent spaces \citep{hafner2019learning,hafner2023mastering,bardes2024v}, but such representations are difficult to inspect or directly edit.
Recent systems instead expose state through code, typed variables, simulators, or programmable scene representations \citep{tang2024worldcoder,lin2026stateplay,cai2026mass,basant2026nvidia,chen2026code,huang2026programmable}.
\method{} differs in that a coding agent acts directly from a goal, selects which object instance changes and when, and records the resulting first-person interaction outcome in a lightweight world state table.

\textbf{Video generation from coarse conditions.}
Video generators can render from structured conditions such as G-buffers, 3D proxies, reprojected geometry, camera parameters, and synthetic trajectory scaffolds \citep{liang2025diffusionrenderer,gomez2026coarse,yu2024viewcrafter,bai2025recammaster,xu2026wonder,chen2026world}.
These conditions primarily encode geometry, layout, and viewpoint.
Building on them, \method{} additionally encodes object state in the conditioning video, so that state changes are expressed directly through changes in the corresponding 3D box.

\section{Methodology}
\label{sec:method}

\begin{figure}[t]
\vspace{-3mm}
\centering
\includegraphics[width=\linewidth]{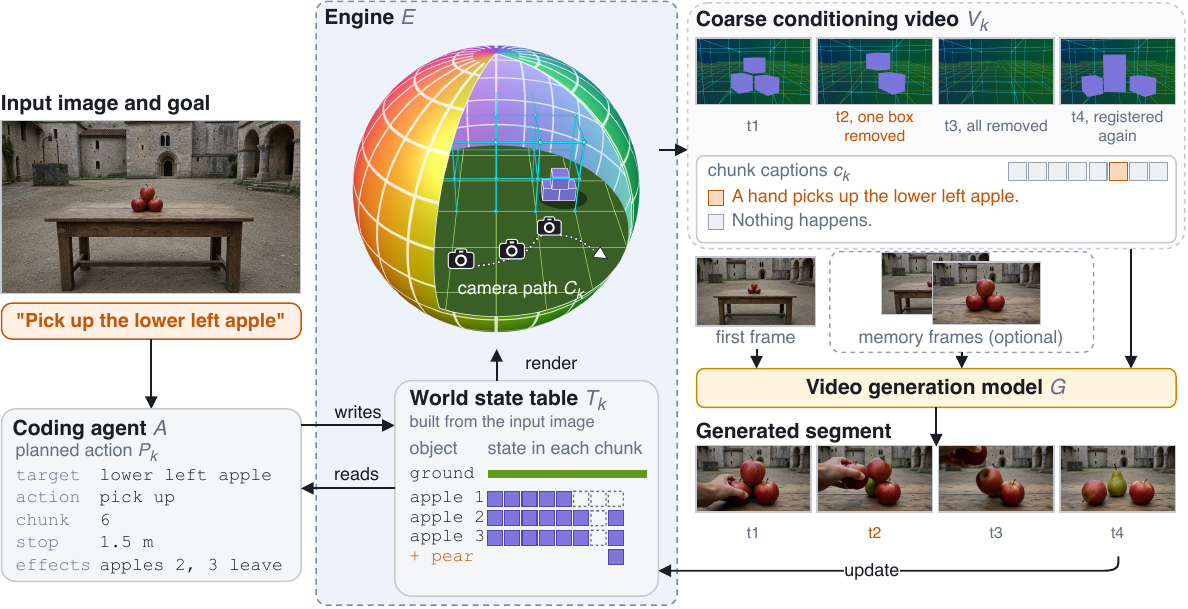}
\vspace{-4mm}
\caption{Pipeline of \method{} for one segment.
Given the input image and a goal, the coding agent $A$ reads the world state table and plans each action $P_k$.
The engine $E$ writes each planned action into the table and renders the result $T_k$ along the camera path $C_k$ as the coarse conditioning video $V_k$ (Section~\ref{sec:state}).
An interaction appears as a change of the object box, such as its removal or recoloring.
Objects that the interaction moves lose their boxes and are registered again from the generated frames once they come to rest, as are objects that it uncovers.
The video model $G$ generates the segment conditioned on the first frame, optional memory frames, $V_k$, and the chunk captions $c_k$ (Section~\ref{sec:videomodel}).
The last frame and the updated table start the next segment.
The sphere is a schematic.
The conditioning and generated frames come from one segment produced by \method.}
\label{fig:pipeline}
\end{figure}

\subsection{The Overall Pipeline}
\label{sec:formulation}
\label{sec:overview}

\method{} is an interactive video world model that starts from an input image $I_0$ and generates an evolving world as a sequence of video segments.
Our goal is to support two key properties of full interaction: \textbf{Open-World Interactivity}, where newly encountered objects can be incorporated into the evolving world and remain interactable, and \textbf{Persistent State}, where the consequences of interaction persist and influence subsequent observations.
To this end, \method{} consists of four components (Fig.~\ref{fig:pipeline}): 1) a coding agent $A$ that interprets goals and plans interactions, 2) an explicit world state table $\mathcal{T}$ that records the evolving objects and their states, 3) a lightweight engine $E$ that executes actions, updates the state, and renders it into visual conditions, and 4) a pretrained video generator $G$ that realizes these conditions as realistic observations.

For segment $k$, a user provides an action input $a_k$, or the coding agent derives the interaction from a higher-level goal.
The agent $A$ reads the previous world state table $\mathcal{T}_{k-1}$ and the first frame $\hat{I}^{(k)}_0$, and produces planned actions $\mathcal{P}_k$.
The engine $E$ executes these actions and writes their outcomes into the table, yielding the updated state $\mathcal{T}_k$.
It then renders $\mathcal{T}_k$ along the camera path $C_k$ into a coarse conditioning video $V_k$ and composes the corresponding captions $c_k$.
Finally, the video generator $G$ produces the segment from the first frame, the conditioning video, and the captions:
\begin{equation}
\begin{aligned}
\mathcal{P}_k
    = A\big(\mathcal{T}_{k-1}, \hat{I}^{(k)}_0, a_k\big), ~~
(\mathcal{T}_k, V_k, c_k)
    = E\big(\mathcal{T}_{k-1}, \mathcal{P}_k\big), ~~
\hat{I}^{(k)}
    \sim G\big(\hat{I}^{(k)}_0, V_k, c_k\big).
\end{aligned}
\label{eq:pipeline}
\end{equation}
After generation, the last frame of the current segment and the updated table $\mathcal{T}_k$ initialize the next segment.
When exploration reveals new objects, the table can be extended from generated observations, so the interactable world is not restricted to the content visible in the initial image.
At the same time, previously modified object states are carried forward through $\mathcal{T}_k$, allowing the consequences of interaction to persist across segments.
This design follows a standard world model \citep{ha2018world,hafner2019learning}.
A standard world model first updates its state through a transition $f$ and then generates an observation through an observation model $p$:
\begin{equation}
s_k = f(s_{k-1}, a_k),
\qquad
o_k \sim p(o \mid s_k).
\label{eq:worldmodel}
\end{equation}

In \method{}, the world state table $\mathcal{T}_k$ plays the role of the explicit state $s_k$, the coding agent $A$ and engine $E$ jointly instantiate the transition $f$, and the video generator $G$ serves as the observation model $p$.
This factorization makes \emph{what happens in the world} explicit and persistently updatable, while leaving \emph{how it is visually realized} to the generative prior of the video model.

\textbf{Explicit World State.} Both Open-World Interactivity and Persistent State are realized through an explicit, evolving world state $\mathcal{T}$.
The state can be extended as new objects are encountered during exploration, while interaction-induced state changes are carried forward across subsequent segments.
Rather than storing visual appearance, $\mathcal{T}$ maintains the interaction-relevant state of the scene, including the walkable ground and, for each object, its semantic identity, 3D position, spatial extent, orientation, presence, and state over time.
\vspace{-0mm}
\subsection{Agentic World State Management}
\label{sec:state}

The world state should contain enough information to specify and preserve interactions, while remaining lightweight enough to construct and extend during generation.
We therefore represent only interaction-relevant geometry and object states, leaving visual appearance and detailed motion to the video model.
At segment $k$, the world state is
\begin{equation}
\mathcal{T}_k
=
\big(
\mathcal{G}_k,\mathcal{O}_k
\big),
\qquad
\mathcal{O}_k
=
\{e_i^{(k)}\}_{i=1}^{M_k},
\end{equation}
where $\mathcal{G}_k$ denotes the walkable ground and $M_k$ is the number of objects currently represented.
Each object entry is $e_i^{(k)} = \big( \ell_i,\mathbf{c}_i,\mathbf{s}_i,\psi_i, \rho_i^{(k)},\sigma_i^{(k)} \big),$ where $\ell_i$ is the semantic label, $\mathbf{c}_i\in\mathbb{R}^3$ and $\mathbf{s}_i\in\mathbb{R}^3$ are the center and extent of its 3D bounding box in world coordinates, $\psi_i$ is its yaw, $\rho_i^{(k)}$ denotes whether the object is present at that pose, and $\sigma_i^{(k)}\in\{0,1,2\}$ is its interaction state.
We use three states since a generated segment rarely shows more than two state changes clearly, though the formulation extends naturally to more states.
State $0$ is the default state, while the semantics of non-default states are specified by the accompanying text.

\paragraph{Grounding Generated Content for Interaction.} To make generated content physically interactable, the system first turns visually generated objects $\mathcal{O}$ into explicit entities that can be localized, referenced, and modified by subsequent actions.
Given the input image, the coding agent identifies the objects relevant to interaction and assigns their initial states.
A table-building tool then uses off-the-shelf perception models to recover the walkable region $\mathcal{G}_0$ and a 3D bounding box for each object, grounding generated content into the structured world state.
Given an action or goal, the agent reads the current table $\mathcal{T}_{k-1}$ and the first frame of the current segment, and grounds the high-level interaction intent into a planned action
\begin{equation}
\mathcal{P}_k
=
\big(
i_k,\alpha_k,\tau_k,d_k,\mathcal{A}_k
\big),
\end{equation}
where $i_k$ denotes the target object instance, $\alpha_k$ the action category, $\tau_k$ the interaction chunk, $d_k$ the camera stop distance, and $\mathcal{A}_k$ the other objects that the interaction changes as a side effect.
This grounding binds each action to a specific object instance, spatial location, and execution time, so that generated content can be directly acted upon and modified.
The agent further decomposes it into a sequence of interactions and orders them according to their state dependencies.

\paragraph{Writing Interaction Outcomes into World State.} Once an interaction has been grounded to a target instance, action, and execution time, the next step is to make its consequence an explicit change to the world.
The engine maps the planned action $\mathcal{P}_k$ to a deterministic state transition,
\begin{equation}
\mathcal{T}_k
=
U(\mathcal{T}_{k-1},\mathcal{P}_k).
\label{eq:state_update}
\end{equation}
Because an interaction changes only part of the world, the update is local to the affected objects:
\begin{equation}
e_i^{(k)}
=
\begin{cases}
\Phi_{\alpha_i}\!\left(e_i^{(k-1)}\right), & i=i_k \text{ or } i\in\mathcal{A}_k,\\
e_i^{(k-1)}, & \text{otherwise},
\end{cases}
\label{eq:object_update}
\end{equation}
where $\alpha_i$ is the change of object $i$ and $\alpha_{i_k}=\alpha_k$, and $\Phi_{\alpha}$ updates the state $\sigma_i$ when the object remains at its pose, or marks its presence $\rho_i$ as absent when it leaves that pose.
Importantly, $\mathcal{T}_k$ records the \emph{result} of the interaction rather than prescribing its full motion trajectory, so a moved object simply loses its box until it comes to rest.
This keeps the state transition explicit and controllable, while leaving the detailed motion of the interaction to the generative prior of the video model.

\paragraph{Growing and Preserving the Interactive World.} The update above specifies how an interaction changes objects already represented in the world state.
A continuously evolving world, however, must also preserve previous changes and incorporate objects that appear only through later exploration.
After each generated segment, newly encountered objects can therefore be recovered from its frames and added to the current object set,
\begin{equation}
\mathcal{O}_k
\leftarrow
\mathcal{O}_k \cup \Delta\mathcal{O}_k,
\label{eq:state_extension}
\end{equation}
so that $M_k$ can grow as exploration unfolds.
Objects marked absent by the interaction are registered again in the same way when they come to rest in view, and every other entry stays as it is.
At the same time, object entries already saved in $\mathcal{T}_k$, including those modified by previous interactions, are carried forward to subsequent segments.
The world state therefore evolves along two complementary dimensions: it \emph{expands} as newly encountered objects become interactable, and it \emph{persists} as the consequences of previous interactions remain part of the world.
Having built how the world state is constructed, updated, and maintained, the last question is how to communicate this structured state to the video generator.
We address this next by rendering it into frame-aligned visual conditions.

\subsection{State-Consistent Video Generation}
\label{sec:videomodel}

Once the world state and interaction logic have been specified, the remaining task is to realize them as realistic visual observations.
The world state determines which object changes, when the change occurs, and what state results, but it does not prescribe the detailed appearance, motion, or interaction dynamics.
We therefore leave these details to the generative prior of the video model.

\textbf{State-Consistent Visual Context .} To make the generated segment both state-consistent and visually coherent with the evolving world, the video generator is conditioned on three complementary sources of context: the current first frame $\hat{I}^{(k)}_0$, the rendered conditioning video $V_k$, and optional memory frames $\mathcal{M}_k$ from earlier segments.
The first frame preserves the local appearance of the current scene, while $V_k$ specifies the camera motion and the object-state changes that should be realized.
Following Wonder \citep{xu2026wonder}, $V_k$ draws an environment sphere and a cube lattice as a camera scaffold, together with the ground grid and the object boxes of the table, whose colors encode the object states (Fig.~\ref{fig:pipeline}).
For longer-range consistency, we retrieve memory frames from previously observed views, but retain only those whose object states agree with the current world state $\mathcal{T}_k$.
This prevents stale visual evidence from contradicting interaction-induced state changes.
The generated segment is modeled:
\begin{equation}
\hat{I}^{(k)}
\sim
G\!\left(
\hat{I}^{(k)}_0,\,
V_k,\,
\mathcal{M}_k,\,
c_k
\right),
\label{eq:video_generation}
\end{equation}
where $c_k$ is the chunk-wise caption of the segment, one caption per chunk that identifies the action in the chunk where it begins.
In this design, the explicit world state determines \emph{what must happen}, while the video model fills in \emph{how it looks and moves}.
To enable it, the generator must be able to interpret the state-aligned conditions defined above and realize them at the correct object and time.

\paragraph{Learning State-Aligned Generation.} To make a pretrained video model natively understand our world-state conditions.
We therefore fine-tune the generator on videos paired with the same conditions used at inference, so that it learns to follow the prescribed camera motion, target object, interaction timing, and resulting object state while retaining its pretrained visual and motion priors.
Because the object-state conditions are {\em category-agnostic}, the learned interaction control can also transfer to object categories unseen during fine-tuning.
Data construction and training details are provided in Section~\ref{sec:experiments} and Appendix~\ref{app:render}.

\section{Experiments}
\label{sec:experiments}
\begin{figure}[t]
\centering
\includegraphics[width=\linewidth]{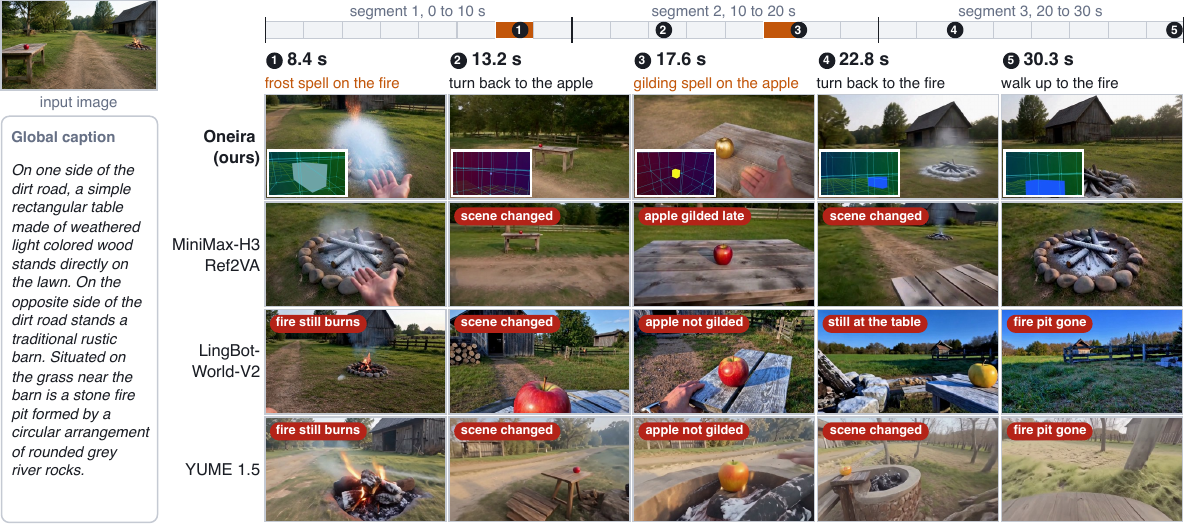}
\caption{Qualitative comparison on a 30-second chain of three segments.
The label above each column gives the action at that moment, orange cells mark the chunks whose captions carry an interaction, and the insets show the conditioning video of \method{}.
Red tags name the failure in a baseline frame.
\method{} controls both interactions at the specified chunks and maintains the changed states when the camera turns back.}
\label{fig:memory_compare}
\vspace{-3mm}
\end{figure}

\subsection{Experimental setup}
\label{sec:setup}

\paragraph{Implementation details.}
We fine-tune the MiniMax-H3 Ref2VA \citep{minimax2026h3} backbone with a LoRA \citep{hu2021lora} of rank \num{128} on the attention projections of the DiT.
The conditioning video is rendered at $448 \times 256$ in both training and inference, while the generated video has $832 \times 480$ pixels.
Training uses \num{32} clips per step and a learning rate of $10^{-4}$ (see details in Appendix~\ref{app:render}).
We use Claude Opus 4.8 as the coding agent, and it builds the world state table with off-the-shelf perception models.
It plans every action in our experiments from the first frame and the action text, and the engine executes the action and renders the conditioning video with OpenGL.

\paragraph{InteractionBench.}
We build InteractionBench to test the interaction and memory abilities of video world models.
It has \num{100} interaction cases and \num{50} memory cases, and GPT-Image-2 generates every first frame.
Each interaction case pairs a first frame with one or two action from our \num{12} action categories.
The interaction cases come in a single-target group of \num{50} and a multi-target group of \num{50}.
Each multi-target case holds five identical objects of one kind, or three identical objects of each of two kinds.
Each memory case is a chain of two segments and lasts \num{20} seconds.
The first segment performs an interaction, and the second segment turns the camera away and returns it to a pose visited after the interaction.

\paragraph{Baselines.} Our baselines are the MiniMax-H3 Ref2VA backbone and three interactive video world models, LingBot-World-V2 \citep{gao2026infinite}, YUME 1.5 \citep{mao2026yume1}, and AlayaWorld \citep{team2026alayaworld}.
These video world models all support interactive events.
We convert the first frame, the camera path, and the action text into the input format of each baseline.
LingBot-World-V2 and AlayaWorld receive our camera path as per-frame poses, while YUME 1.5 and MiniMax-H3 Ref2VA take no poses and receive the camera motion as text per chunk.
In the memory part, MiniMax-H3 Ref2VA receives the same number of memory frames as \method{}.

\paragraph{Metrics.}
On the interaction part, we define three scores, Interaction, Timing, and Target.
Interaction asks whether the specified action takes place, Timing evaluates whether any action begins in the specified chunk, and Target evaluates whether the changed instance is the target object.
We score Interaction and Timing with Gemini 3.7 Flash \citep{team2023gemini} as the VLM judge on the generated frames.
For Target, we follow our data pipeline to build the world state table on the generated video and check which instance changes.
On the memory part, PSNR and SSIM between the visit and revisit frames at the same pose measure global consistency.
State asks the judge whether the target object keeps its changed state.
For visual quality, we report eight VBench \citep{huang2024vbench} dimensions on all \num{150} cases (see details in Appendix~\ref{app:eval}).

\begin{figure}[t]
\centering
\includegraphics[width=\linewidth]{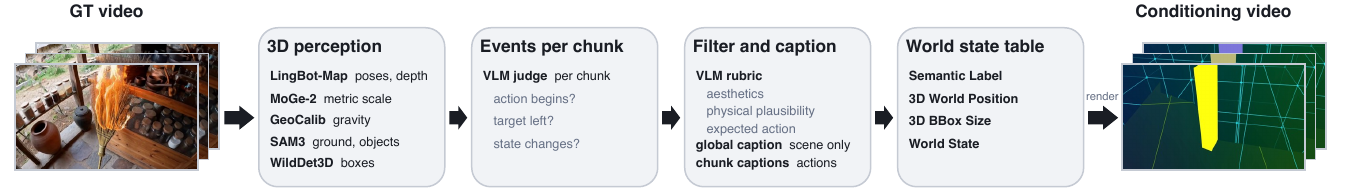}
\vspace{-3mm}
\caption{Data pipeline.
We recover the camera path, the walkable ground, and the object boxes from the GT video, and a VLM judge annotates the state change.
Gates keep the clean clips, fixed rules assemble the world state table, and the engine renders the conditioning video from it.}
\label{fig:data_pipeline}
\vspace{-4mm}
\end{figure}

\subsection{Data pipeline}
\label{sec:data}
\label{sec:datasources}
We build an annotation pipeline that turns videos into world state tables and renders their conditioning videos, yielding a corpus of about \num{10}K ten-second clips.
Fig.~\ref{fig:data_pipeline} summarizes the pipeline.
More details are provided in Appendix~\ref{app:data}.

\paragraph{Sources.} (1) We synthesize first-person videos of \num{12} state-changing actions with proprietary video models.
(2) We incorporate navigation videos from two public datasets, the real walking videos of Sekai \citep{li2025sekai} and the game footage of ABot \citep{jiang2026abot}.
About \num{60}\% of the clips are interaction and \num{40}\% navigation, at $832 \times 480$ and \num{24} fps, and about \num{1}K clips of both kinds come with memory frames of the same place.

\paragraph{3D perception and events.}
LingBot-Map \citep{chen2026geometric} estimates the camera poses and depth, MoGe-2 \citep{wang2025moge2accuratemonoculargeometry} the metric scale, GeoCalib \citep{veicht2024geocalib} the gravity direction, SAM3 \citep{carion2026sam} the masks of the ground and the named objects, and WildDet3D \citep{huang2026wilddet3d} lifts each object to a 3D box in world coordinates.
Moreover, we use Gemini 3.7 Flash to determine the start, duration, and end of each action as its action label.

\paragraph{Filtering and captioning.}
For each video, Gemini 3.7 Flash writes a checklist for physical plausibility and the correctness of the action, and then judges the video against it.
We also check for scene cuts and aesthetics, and we discard every clip that fails.
Following LingBot-World \citep{gao2026infinite}, the global caption describes the scene without any camera motion or interaction, and these dynamics come only from the conditioning video and the chunk captions.

\paragraph{Table and rendering.}
To keep training and inference consistent, we build a world state table for each clean clip and render the conditioning video from it.
The table records each object by its semantic label, one 3D box in world coordinates for its position and size, and its state in each chunk.
The engine renders the video along the estimated camera path and composes the chunk captions with the same code as at inference.

\subsection{Quantitative results}
\label{sec:realization}
\label{sec:held}

\begin{table}[t]
\vspace{-2mm}
\caption{Results on InteractionBench.
Interaction, Timing, and Target are pass rates on the \num{100} interaction cases.
PSNR in dB and SSIM compare the visit and revisit frames, and State is the pass rate on the \num{50} memory chains.
VBench reports eight dimensions.
Bold marks the best entry based on unrounded scores.}
\label{tab:main}
\centering
\small
\setlength{\tabcolsep}{3pt}
\resizebox{\linewidth}{!}{%
\begin{tabular}{@{}lcccccccccccccc@{}}
\toprule
& \multicolumn{3}{c}{Interaction part $\uparrow$} & \multicolumn{3}{c}{Memory part $\uparrow$} & \multicolumn{8}{c}{VBench $\uparrow$} \\
\cmidrule(lr){2-4} \cmidrule(lr){5-7} \cmidrule(l){8-15}
Method & Interaction & Timing & Target & PSNR & SSIM & State & SC & BC & MS & TF & DD & AQ & IQ & OC \\
\midrule
MiniMax-H3 Ref2VA & 0.30 & 0.57 & 0.53 & 10.96 & 0.34 & 0.68 & 0.83 & 0.90 & 0.99 & 0.97 & \textbf{1.00} & 0.53 & 0.73 & 0.19 \\
LingBot-World-V2 & 0.49 & 0.40 & 0.39 & 9.08 & 0.27 & 0.44 & 0.79 & 0.87 & 0.97 & \textbf{0.98} & 0.97 & 0.54 & 0.72 & 0.20 \\
YUME 1.5 & 0.05 & 0.43 & 0.22 & 10.09 & 0.35 & 0.41 & 0.83 & 0.89 & 0.99 & 0.97 & 0.95 & 0.52 & \textbf{0.74} & 0.20 \\
AlayaWorld & 0.08 & 0.16 & 0.26 & 9.19 & 0.30 & 0.39 & 0.79 & 0.88 & 0.99 & 0.95 & 0.65 & 0.49 & 0.69 & 0.18 \\
\midrule
\method{} & \textbf{0.78} & \textbf{0.76} & \textbf{0.82} & \textbf{21.11} & \textbf{0.64} & \textbf{0.80} & \textbf{0.84} & \textbf{0.90} & \textbf{0.99} & 0.97 & \textbf{1.00} & \textbf{0.54} & 0.73 & \textbf{0.22} \\
\bottomrule
\end{tabular}}
\end{table}

In Table~\ref{tab:main}, we present the results on InteractionBench, evaluating every method with the protocol of Section~\ref{sec:setup}.
On the interaction part, \method{} reaches the highest rate on all three scores.
\textbf{\method{} outperforms the best baseline by \num{0.29} in Interaction, \num{0.19} in Timing, and \num{0.29} in Target.}
The baselines often start some action in the specified chunk, since Timing exceeds Interaction for three of them, but the change is the wrong one or falls on the wrong instance.
These results show the advantage of our representation for world modeling.
The world state table encodes the world explicitly and efficiently in an abstract, structured form, and this form helps the model keep the correct time and target of each interaction.
\textbf{For the memory part, \method{} improves PSNR by \num{93}\%, SSIM by \num{83}\%, and State by \num{18}\% over the second-best method.}
MiniMax-H3 Ref2VA keeps the changed state with the memory frames and reaches a State of \num{0.68}, but it has no camera control, so its revisit frame comes from a different pose and its PSNR and SSIM stay low.
The other baselines cannot keep an updated world state for long once the camera turns away, so the scene and the changed state drift by the revisit.
By contrast, \method{} maintains the updated state across segments with the world state table and the memory frames, and the camera returns to the same pose.
On VBench, \method{} improves subject consistency and overall consistency by 2.14\% and 8.58\%, respectively, over the strongest baseline for each metric, while achieving the highest motion smoothness, supporting coherent, smooth, and text-aligned video generation under explicit state conditioning.

\subsection{Qualitative results}
\label{sec:qualitative}
Fig.~\ref{fig:memory_compare} demonstrates a qualitative comparison between our method and the baselines on a chain of three segments over \num{30} seconds.
\method{} puts out the campfire and gilds the apple inside the specified chunks, and both states are still in place when the camera turns back.
MiniMax-H3 Ref2VA keeps the appearance of both changed objects, but it gilds the apple about one chunk late and its scene changes when the camera turns back.
LingBot-World-V2 and YUME 1.5 do not put out the fire, and their scenes change after the first turn.
In \method{}, the memory frames carry the appearance of the changed objects, the world state table fixes their positions, and the conditioning video fixes the camera turn (more in Appendix~\ref{app:qual}).

\subsection{Ablation studies}
\label{sec:ablations}
\begin{table}[t]
\vspace{-4mm}
\caption{Ablation of the conditions on the interaction part of InteractionBench.
RotErr in degrees and TransErr are Umeyama-aligned Pi3X camera metrics.}
\label{tab:ablation}
\centering
\small
\begin{tabular}{@{}lccccc@{}}
\toprule
Variant & Interaction $\uparrow$ & Timing $\uparrow$ & Target $\uparrow$ & RotErr $\downarrow$ & TransErr $\downarrow$ \\
\midrule
Ours w/o sphere & 0.63 & 0.69 & 0.62 & 3.32 & 0.20 \\
Ours w/o boxes & 0.61 & 0.74 & 0.61 & 2.78 & 0.15 \\
Ours w/o chunk captions & 0.73 & 0.66 & 0.80 & 2.71 & 0.15 \\
\midrule
Ours (full) & \textbf{0.78} & \textbf{0.76} & \textbf{0.82} & \textbf{2.60} & \textbf{0.13} \\
\bottomrule
\end{tabular}
\vspace{-3mm}
\end{table}

We perform an ablation study to evaluate the contributions of the key components of our framework.
We train three variants, each with one condition of the video model removed, and evaluate them on the interaction part of InteractionBench.
Ours w/o sphere removes the environment sphere and the cube lattice from the conditioning video, to observe their effect on camera motion.
Ours w/o boxes removes the object boxes, to observe their effect on the accuracy of the interaction.
Ours w/o chunk captions removes the chunk captions and states the time of the action in the global caption.
RotErr and TransErr follow the camera accuracy protocol of SANA-WM \citep{zhu2026sana} (see details in Appendix~\ref{app:eval}).
Table~\ref{tab:ablation} shows the quantitative comparison between our full method and these variants.
Ours w/o sphere loses most of the cues to the camera motion, so the motion becomes ambiguous and camera control drops sharply.
The wrong camera motion also lowers its interaction scores.
Ours w/o boxes lacks the instance-level condition, so the model has to pick the target instance from the action text alone and often changes a wrong object.
Ours w/o chunk captions shows that a prompt alone does not place the action accurately in time, since the global caption states the time only as text and the model has to guess the chunk.
\vspace{-1mm}
\section{Conclusion, Limitations, \& Future Work}
\vspace{-1mm}
In this paper, we proposed \method{}, a generative video world model that allows full interaction, with the ability to maintain a world state which maintains the influence of interactions on subsequent observations and interactions.
Experiments show we make it successful in open-world interactivity and state continuity.
The newly generated objects can be interacted with, and the changes to those objects can be reflected in future observations.
One of the limitations is that in \method{}, the current action space is still discrete.
Future work could extend the framework to more fine-grained physical interactions and continuous-control settings, such as precise manipulation and robot control.
The physical realism of complex interactions is likewise limited by the learned dynamics of the video model rather than an explicit physics simulator.
While the explicit world state improves the persistence of interaction outcomes, long-horizon visual consistency is still bounded by the underlying video generator.
Therefore, we plan to design more data to improve the pretrained video models.

\subsection*{AI use statement}
AI-assisted tools were used for language polishing of the manuscript and for code development.
All scientific content, methodology, experiments, analysis, and conclusions were developed by the authors, who take full responsibility for all content in the paper.

\subsection*{Ethics statement}
This work studies interactive video world models and does not involve human subjects or the collection of personally identifiable information.
The models developed in this work may inherit biases and limitations from their pretrained components and training data.
More capable generative world models could also potentially be misused to generate misleading or synthetic visual content.
We therefore emphasize responsible use of the proposed system for research and simulation purposes and encourage appropriate safeguards when deploying such models in real-world applications.

\subsection*{Reproducibility statement}
We provide detailed descriptions of the model architecture, world-state representation, training procedure, conditioning mechanism, evaluation protocol, and experimental settings in the main paper and supplementary material.
We report the datasets, pretrained models, and key hyperparameters used in our experiments.
Code and additional implementation details will be released to facilitate reproduction of the reported results.

\bibliography{iclr2027_conference}
\bibliographystyle{iclr2027_conference}

\appendix
\section{Further related works}
\label{app:relatedwork}
\method{} consists of a coding agent that maintains an explicit world state and a video model that renders observations from it.
We organize the related work around these two components and the interface between them.

\paragraph{Interactive video world models.}
Interactive video world models predict how an environment evolves under the actions of an agent.
Game world models condition generation on player inputs \citep{valevski2025diffusion,decart2024oasis,alonso2024diffusion,guo2025mineworld,he2025matrix,wang2026matrix}, and camera-controlled systems extend exploration to minute-long rollouts that remain consistent upon revisits \citep{sun2025worldplay,hong2025relic,xu2026wonder,zhu2026sana,xiao2025worldmem,yu2025context}.
UniSim \citep{yang2023learning} conditions generation on robot actions and language instructions, and DreamGen \citep{jang2025dreamgen} generates robot data for policy learning.
Genie~3 and related systems support text-triggered world events \citep{parker2025genie,xiang2025pan,mao2026yume1}, ActWorld \citep{xiong2026actworld} adds action-aware memory to preserve modified object states, and LingBot-World~2.0 \citep{gao2026infinite} places agents that issue semantic actions inside the rollout.
These advances strengthen control, memory, and agentic interaction around a learned visual rollout.
Their central modeling object, however, remains the generated observation sequence.
Interaction outcomes are represented only implicitly in visual history, latent context, or memory tokens, and they cannot be verified against an explicit specification.
\method{} specifies the outcome of each interaction in a world state table before generation and uses the video model to render it.

\paragraph{Abstract and explicit world state.}
A second line of work models the effect of actions in a space other than pixels.
Model-based reinforcement learning learns world dynamics in a latent space \citep{hafner2019learning,hafner2023mastering}, and V-JEPA predicts video in a representation space \citep{bardes2024v}.
These learned abstractions are hard to interpret, and an intended outcome can be neither written into them nor decoded from them.
Recent work represents the state explicitly.
Program-based world models express dynamics as code that an agent can inspect and revise \citep{tang2024worldcoder,lehrach2026code}.
For interactive video, StatePlay \citep{lin2026stateplay} jointly predicts observations and game-state variables, and MASS \citep{cai2026mass} advances a typed state with a learned logic engine.
OmniDreams \citep{basant2026nvidia} conditions closed-loop driving simulation on a rendered abstract state of lane lines and road actors.
This state is derived from a traffic simulator and driving logs, and its changes are limited to the motion of road actors.
Concurrently, Code World Model \citep{chen2026code} uses a coding agent to maintain world state through code and compiles the state into a proxy video.
Programmable World Model \citep{huang2026programmable} compiles state-augmented 3D bounding boxes into controls for a video renderer.
In both systems, the agent programs the entity states and the transition rules of a playable world, and a player acts in it.
In \method, the rules are fixed in a lightweight engine, and the coding agent itself acts from a goal, deciding which object instance changes and when.
It targets the state changes that first-person interactions cause to objects, and encodes each change as the removal or recoloring of a 3D box in the conditioning video.

\paragraph{Video generation from coarse conditions.}
Video models can serve as renderers for structured conditions of varying explicitness.
G-buffers \citep{liang2025diffusionrenderer,huang2026generative} and coarse 3D proxies \citep{gomez2026coarse} constrain geometry and layout, re-rendered point clouds anchor camera control to observed geometry \citep{yu2024viewcrafter,team2026inspatio}, and camera parameters can be injected as embeddings \citep{bai2025recammaster,wang2026bullettime,he2025cameractrl}.
Wonder \citep{xu2026wonder} renders a synthetic lattice and an environment map along the trajectory, so that the model interprets camera motion directly as visual evidence.
At the other end of this spectrum, a white-box world built in a 3D engine drives a frozen video model frame by frame \citep{chen2026world}.
These conditions specify geometry, layout, and viewpoint, and they carry no explicit notion of object state.
\method{} follows the camera scaffold of Wonder and encodes the state of each object in the same conditioning video as a colored box.
A state change appears as the removal or recoloring of that box.

\section{Training and rendering details}
\label{app:render}

We fine-tune the backbone with fully sharded data parallelism in its hybrid form over \num{32} GPUs, one clip per GPU, so a step sees \num{32} clips.
Each clip has \num{243} frames encoded into \num{72} latent frames, \num{9} per chunk, and the first frame and the conditioning video enter through the reference inputs of the backbone, packed ahead of the video latents with its official packed-sequence builder.
The LoRA has rank \num{128} on the attention projections and the learning rate is $10^{-4}$.
In half of the training steps we fix the first frame as at inference, and in the other half we leave it free, so the model keeps its generative ability.

The global caption and the eight chunk captions are concatenated into one text sequence, and a block-causal mask lets the video tokens of chunk $c$ attend to the global caption and to the captions of chunks up to $c$, following the chunk-wise captions of LingBot-World-V2 \citep{gao2026infinite}.
A chunk caption contains an action sentence and an outcome sentence, such as \texttt{A hand picks up the red apple.} followed by \texttt{The apple has left the table.}
The global caption contains one fixed passage shared by every sample, which explains each element of the conditioning video, including the meaning of the three colors and of a disappearing box.

The engine draws each object entry as a solid box at $(\mathbf{c}_i, \mathbf{s}_i, \psi_i)$ over the frames in which its presence $\rho_i$ holds.
The lattice has a pitch of \num{1.5} m, two levels, and a radius of \num{6.6} m with distance fade, and the ground grid has the same pitch and is drawn at an alpha of \num{0.12}.
Rasterization takes about \num{9} s per clip, mostly spent on reading pixels back from the GPU and encoding the video.
At inference, the first frame is kept fixed, since after every denoising step the first latent frame is reset to $\alpha \mathbf{z}_0 + (1-\alpha) \boldsymbol{\epsilon}$ with $\alpha = 0.999$, which matches the image-to-video loss used in training.

\section{Data details}
\label{app:data}

\subsection{Annotation}
\label{sec:annotation}
\label{sec:timings}
For each training video, LingBot-Map \citep{chen2026geometric} estimates the camera poses and depth, and GeoCalib \citep{veicht2024geocalib} estimates the gravity direction.
MoGe-2 \citep{wang2025moge2accuratemonoculargeometry} provides the metric scale.
SAM3 \citep{carion2026sam} segments the named objects and the walkable ground, and WildDet3D \citep{huang2026wilddet3d} lifts the objects to 3D bounding boxes in world coordinates.
We fit the ground plane to the ground points with RANSAC constrained by the gravity direction and grow its walkable region across the clip.
For each chunk, Gemini 3.7 Flash reports whether an action begins, whether the target has left its position, and whether a state change is under way.
From these answers we assemble the event chunk, the transition window, and the chunk captions with fixed rules, so the captions and the conditioning video are consistent by construction.

\subsection{Composition of the training set}
\label{app:composition}
\begin{figure}[h]
\centering
\includegraphics[width=\linewidth]{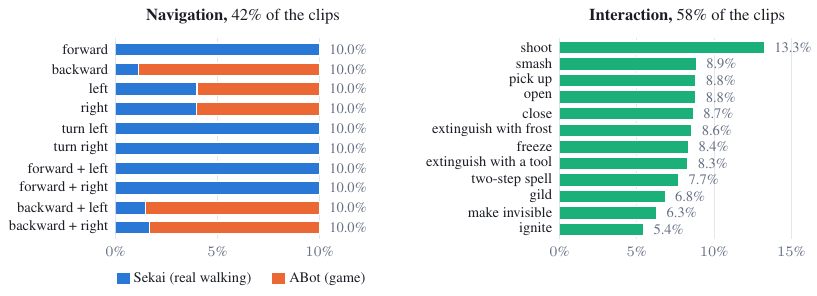}
\caption{Composition of the training set.
Left, the share of navigation clips per category of camera motion and their source.
Right, the share of interaction clips per action category.}
\label{fig:datadist}
\end{figure}

The training set holds about \num{5800} interaction clips and about \num{4200} navigation clips.
When we synthesize the interaction clips with the proprietary video models, each prompt combines one of the \num{12} action categories with a target object and a scene, and we balance the prompts over the actions, the scenes, and the chunk in which the interaction begins.
The model therefore sees interactions early, midway, and late in a clip.
Fig.~\ref{fig:datadist} shows the share of clips per action category.

\subsection{Navigation videos}
\label{app:navdata}
The navigation pool holds about \num{4200} clips and supplies camera motion in scenes where nothing happens, real walking videos from Sekai \citep{li2025sekai} and first-person game captures with recorded camera poses \citep{jiang2026abot}.
Backward and sideways motion is rare in real walking videos, so most of these clips come from the game footage.
We filter the pool in three steps.
TransNetV2 \citep{soucek2024transnet} removes clips with a shot change, DOVER \citep{wu2023exploring} scores aesthetic and technical quality and removes the low end, and the optical flow of UniMatch \citep{xu2023unifying} removes clips with too little or too much motion.
Gemini 3.7 Flash then rejects clips with glare, darkness, overlays, blank frames, a dominant foreground object, or a focus or zoom shift, and it counts the people and vehicles in each clip, so we keep clips with few of them, a straight path, and enough visible ground.
Finally, we estimate the camera path with LingBot-Map and MoGe-2 as in Section~\ref{sec:data} and reject clips whose camera height or displacement falls outside a plausible range.
In the clips that remain, the table also contains the people and vehicles, whose boxes follow fitted trajectories, and every chunk caption is the placeholder \texttt{Nothing happens.}

\section{Evaluation protocol}
\label{app:eval}

\begin{figure}[t]
\centering
\includegraphics[width=\linewidth]{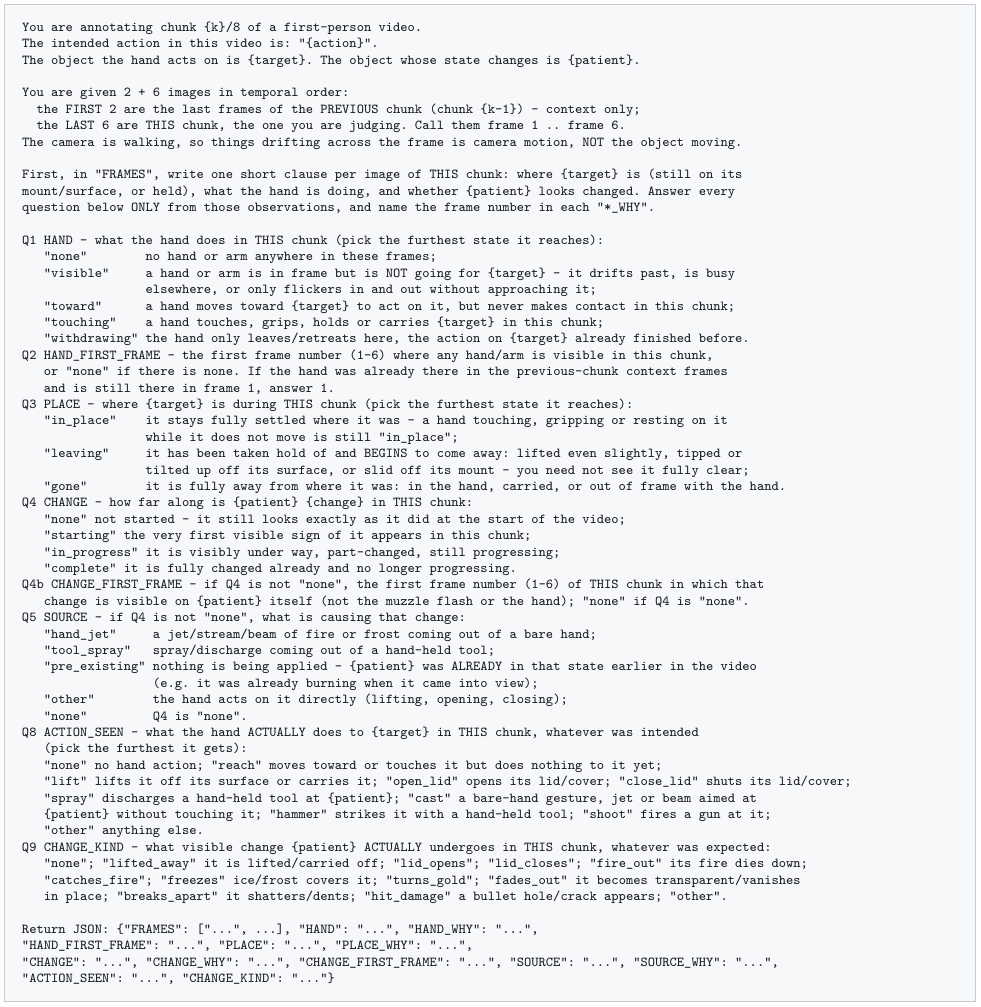}
\caption{The prompt of the VLM judge for one chunk.
Placeholders in braces are filled per case, and the questions on hand-held tools and two-step spells are omitted.}
\label{fig:judge_prompt}
\end{figure}

Every method generates one video per case from the same first frame, camera input, and action text.
We cut each video into eight chunks of equal duration, so a chunk covers the same span of the case for every method regardless of its frame rate.
Our videos have \num{243} frames at \num{24} fps and $832 \times 480$ pixels.

\noindent\textbf{Interaction and Timing.}
Gemini 3.7 Flash reads six frames sampled uniformly from each chunk, together with the last two frames of the chunk before it as context, and the action text.
For each chunk, it reports the hand action it sees and the kind of change it sees, each from a fixed list, and whether the specified change has taken place, with the prompt shown in Fig.~\ref{fig:judge_prompt}.
Interaction counts a case as correct if in some chunk the change and the hand action match the ones expected for the action category and a hand performs the action.
Timing counts a case as correct if the first chunk in which the judge sees an action begin is the specified chunk.

\noindent\textbf{Target.}
We run the 3D perception of our data pipeline on each generated video and build its world state table, which gives every object instance a 3D box in the world coordinates anchored at the first frame and a state per chunk.
The specification fixes the target instance by its box in the table built from the first frame, so we match the instance that changes its state in the generated video to that box.
Target counts a case as correct if the changed instance is the specified one, whatever the change is.
For the baselines, we name the specific target object in the action text.

\noindent\textbf{Memory.}
Each chain fixes one visit frame in the first segment and one revisit frame in the second at the same camera pose from the specification.
PSNR and SSIM are computed between these two frames.
For State, the judge sees the revisit frame and the action text and answers whether the target object is in its changed state.

\noindent\textbf{Visual quality.}
We run VBench \citep{huang2024vbench} in custom-input mode on all \num{150} videos and report subject consistency (SC), background consistency (BC), motion smoothness (MS), temporal flickering (TF), dynamic degree (DD), aesthetic quality (AQ), imaging quality (IQ), and overall consistency (OC).

\noindent\textbf{Camera accuracy.}
We estimate the camera poses of each generated video with Pi3X \citep{wang2025pi}, relativize the specified path to the first frame, and align the estimate to it with a Umeyama Sim(3) transform.
Let $\mathbf{P}_t=[\mathbf{R}_t|\mathbf{t}_t]$ be the specified pose and $\widetilde{\mathbf{P}}_t=[\widetilde{\mathbf{R}}_t|\widetilde{\mathbf{t}}_t]$ the aligned estimate for frame $t$.
We report
\begin{equation}
\mathrm{RotErr}=\frac{1}{T}\sum_{t=1}^{T}\frac{180}{\pi}
\cos^{-1}\!\left(\operatorname{clip}\!\left(\frac{\operatorname{tr}(\mathbf{R}_t^\top\widetilde{\mathbf{R}}_t)-1}{2},-1,1\right)\right),
\qquad
\mathrm{TransErr}=\frac{1}{T}\sum_{t=1}^{T}\|\mathbf{t}_t-\widetilde{\mathbf{t}}_t\|_2,
\end{equation}
where RotErr is in degrees and both are lower-is-better.

\section{Additional qualitative results}
\label{app:qual}

\begin{figure}[h]
\centering
\includegraphics[width=\linewidth]{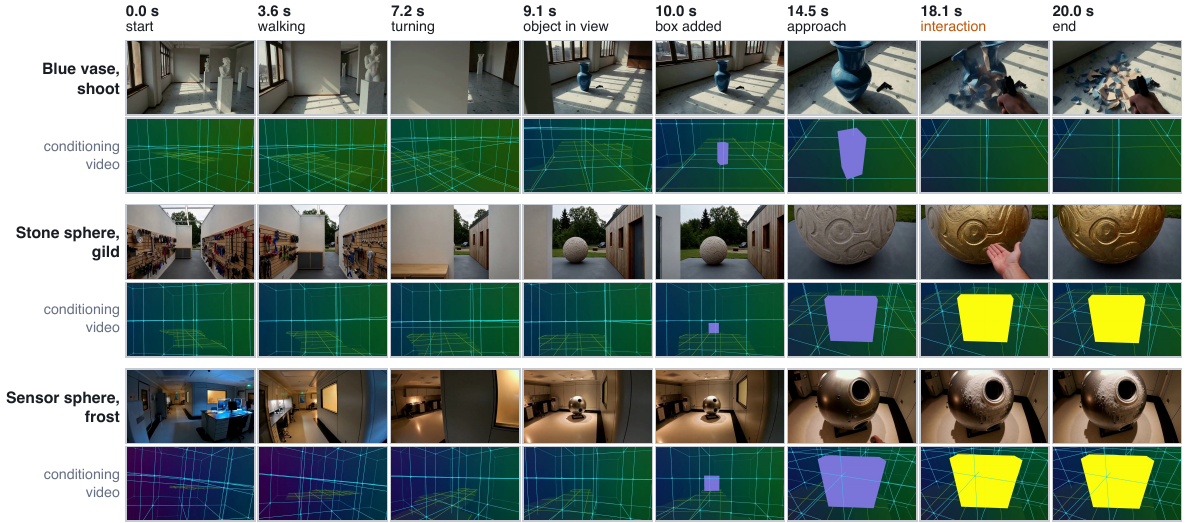}
\caption{Interaction with unseen objects.
The target is absent from the first frame and comes into view while the camera explores in the first segment.
The coding agent adds it to the world state table before the second segment, so its box appears in the conditioning video from then on, and a hand interacts with it.
The row under each chain is its conditioning video.}
\label{fig:qual_unseen}
\end{figure}

\begin{figure}[p]
\centering
\includegraphics[width=\linewidth]{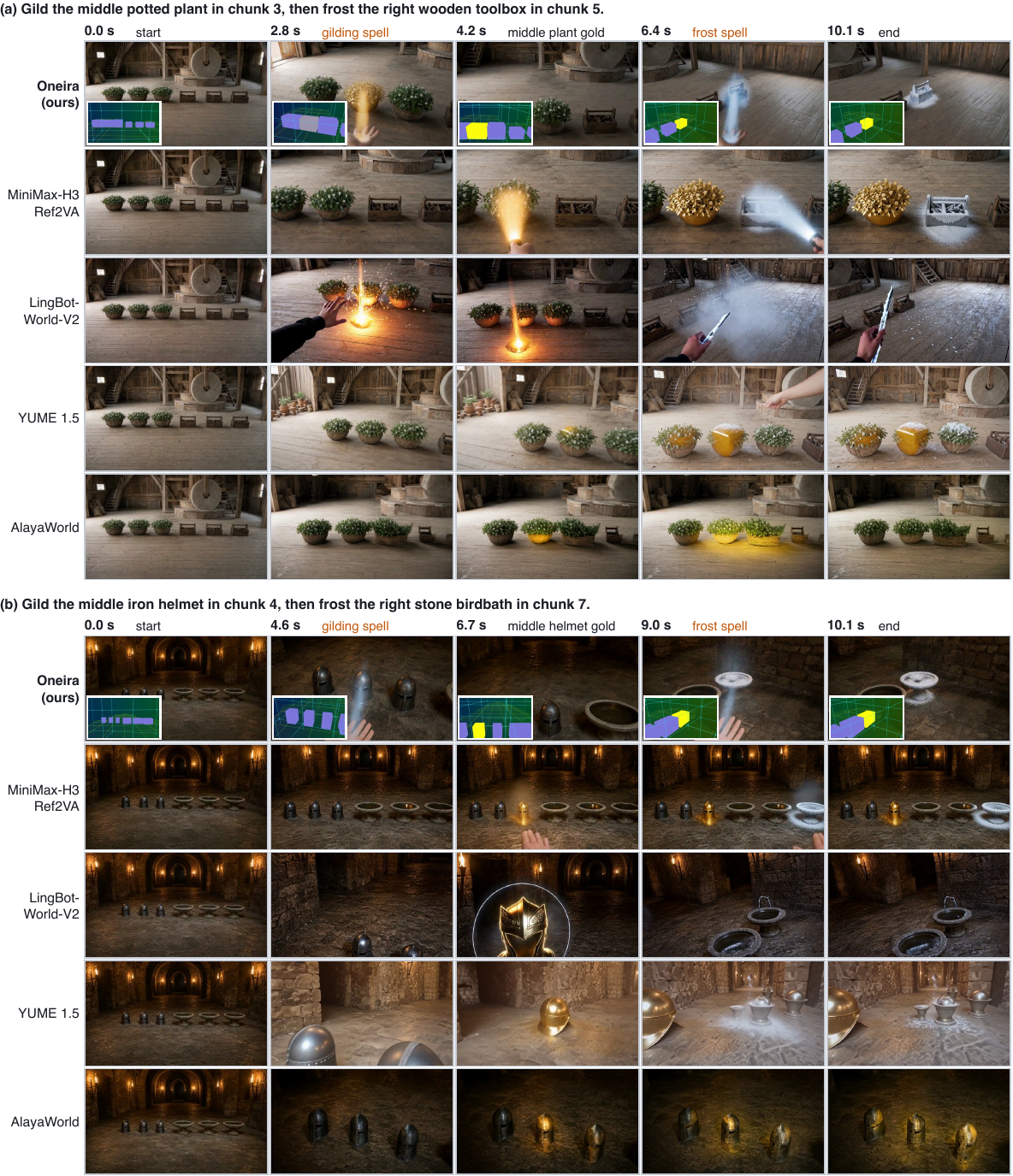}
\caption{Interaction with identical objects.
Each scene holds two kinds of objects, three identical objects of each kind, and two chunk captions name one object each by its place in the row.
Insets show the conditioning video of \method{}, and orange labels mark the interaction columns.}
\label{fig:qual_identical}
\end{figure}

Fig.~\ref{fig:qual_unseen} shows three chains in which the target is absent from the first frame.
The video model generates the object while the camera explores in the first segment, the coding agent registers it in the world state table, and the interaction of the second segment lands on it.
Fig.~\ref{fig:qual_identical} shows two cases of the multi-target group, each with two interactions on two kinds of objects.
\method{} changes the named object in both cases and leaves its neighbours unchanged, while the baselines change a wrong object or several objects, and two of them lose the layout of the row.

\end{document}